\documentclass[10pt,journal,compsoc]{IEEEtran}

\ifCLASSOPTIONcompsoc
  \usepackage[nocompress]{cite}
\else
  \usepackage{cite}
\fi

\usepackage[pdftex]{graphicx}
\graphicspath{{./figs/}}
\DeclareGraphicsExtensions{.pdf,.jpg,.png}
\usepackage{cite}
\usepackage[cmex10]{amsmath}
\usepackage{amssymb}
\usepackage{graphicx}
\usepackage{cases}
\usepackage{mdwmath}
\usepackage{array}
\usepackage{url}
\usepackage[ruled,vlined]{algorithm2e}
\usepackage{booktabs}
\usepackage{threeparttable}
\usepackage{color}
\usepackage{caption}
\usepackage{pifont}
\usepackage{wasysym}
\usepackage{amssymb}
\usepackage{multirow}
\usepackage{makecell}
\usepackage[colorlinks,linkcolor=black]{hyperref}    
\usepackage{marvosym}
\usepackage{lipsum}
\usepackage[caption=false,font=footnotesize]{subfig}
\usepackage{amsmath}
\usepackage{xspace}
\usepackage{bm}
\usepackage{amsthm,amsmath,amssymb}
\usepackage{mathrsfs}
\usepackage{mathrsfs}
\usepackage{multicol} 
\usepackage{hyperref}
\usepackage{tabularx}
\usepackage{longtable}
\usepackage{lipsum}
\usepackage{tikz}
\usetikzlibrary{mindmap,trees}
\usepackage{verbatim}

\hypersetup{
    colorlinks=true,             
    linkcolor=black,             
    filecolor=black,             
    urlcolor=magenta,              
    citecolor=blue,             
    pdftitle={Overleaf Example},
    pdfpagemode=FullScreen,
}

\usepackage{hyperref}
\usepackage{overpic}

\usepackage{zhlipsum}
\usepackage{xcolor}

\usepackage[absolute,overlay]{textpos}

\makeatletter

\newcommand{\Rmnum}[1]{\expandafter\@slowromancap\romannumeral #1@}
\makeatother
\usepackage{graphicx}
\usepackage{float}
\usepackage{diagbox}

\newcommand{\ie}{\textit{i}.\textit{e}., }
\newcommand{\eg}{\textit{e}.\textit{g}., }
\def\etal{{\em et al.}}

\usepackage{colortbl}  
\usepackage{xcolor}
\usepackage{array}   
\usepackage{balance}

\begin{document}

\title{Awesome Multi-modal Object Tracking}

\author{Chunhui~Zhang,\IEEEmembership{~Student Member,~IEEE},
        Li~Liu$^*$,\IEEEmembership{~Member,~IEEE}, Hao Wen, Xi Zhou, Yanfeng Wang
\IEEEcompsocitemizethanks{
\IEEEcompsocthanksitem Chunhui~Zhang is with the Cooperative Medianet Innovation Center, Shanghai Jiao Tong University, Shanghai 200240, China and the Hong Kong University of Science and Technology (Guangzhou), Guangzhou 511458, China and also with the CloudWalk Technology Co., Ltd, 201203, China. Email: chunhui.zhang@sjtu.edu.cn.\protect

\IEEEcompsocthanksitem Li~Liu is with the Hong Kong University of Science and Technology (Guangzhou), Guangzhou, 511458, China. E-mail: avrillliu@hkust-gz.edu.cn.\protect

\IEEEcompsocthanksitem Hao Wen, and Xi Zhou are with the CloudWalk Technology Co., Ltd, 201203, China. E-mails: wenhao@cloudwalk.com, zhouxi@cloudwalk.cn.\protect

\IEEEcompsocthanksitem Yanfeng Wang is with the Cooperative Medianet Innovation Center, Shanghai Jiao Tong University, Shanghai, 200240, China and the Shanghai AI Laboratory. E-mail: wangyanfeng@sjtu.edu.cn.\protect

}

\thanks{$^{*}$ Corresponding author.}
\thanks{This work was done at the Hong Kong University of Science and Technology (Guangzhou).}

}

\markboth{A Report for Awesome MMOT Project, Start Date: May 20, 2024}%
{Shell \MakeLowercase{\textit{et al.}}: Bare Advanced Demo of IEEEtran.cls for IEEE Computer Society Journals}

\IEEEtitleabstractindextext{%
\begin{abstract}
Multi-modal object tracking (MMOT) is an emerging field that combines data from various modalities, \eg vision (RGB), depth, thermal infrared, event, language, and audio, to estimate the state of an arbitrary object in a video sequence. It is of great significance for many applications such as autonomous driving and intelligent surveillance. In recent years, MMOT has received more and more attention. However, existing MMOT algorithms mainly focus on two modalities (\eg RGB+depth, RGB+thermal infrared, and RGB+language). To leverage more modalities, some recent efforts have been made to learn a unified visual object tracking model for any modality. Additionally, some large-scale multi-modal tracking benchmarks have been established by simultaneously providing more than two modalities, such as vision-language-audio (\eg WebUAV-3M) and vision-depth-language (\eg UniMod1K). To track the latest progress in MMOT, we conduct a comprehensive investigation in this report. Specifically, we first divide existing MMOT tasks into five main categories, \ie RGBL tracking, RGBE tracking, RGBD tracking, RGBT tracking, and miscellaneous (RGB+X), where X can be any modality, such as language, depth, and event. Then, we analyze and summarize each MMOT task, focusing on widely used datasets and mainstream tracking algorithms based on their technical paradigms (\eg self-supervised learning, prompt learning, knowledge distillation, generative models, and state space models). Finally, we maintain a continuously updated paper list for MMOT at {\color{magenta}https://github.com/983632847/Awesome-Multimodal-Object-Tracking}.

\end{abstract}

}

\maketitle
\IEEEdisplaynontitleabstractindextext
\IEEEpeerreviewmaketitle

\section{Background and Motivation}
\label{sec:motivation}

Although RGB-based object tracking methods have made significant progress over the past decade, they still cannot achieve precise and robust tracking in some complex situations, such as lighting changes, fast motion, occlusion, and appearance variations. To address this issue, some researchers have proposed the task of multi-modal object tracking (MMOT)~\cite{chenglong2023mmot}, which introduces additional modalities such as thermal infrared, depth, event, and language modalities to compensate for the shortcomings of the RGB modality under adverse weather conditions, occlusions, rapid motion, and appearance ambiguity. 

MMOT can leverage the complementary advantages of RGB and other modalities to achieve more robust target location in videos, which has garnered increasing research interest and attention. This initially inspired us to conduct an investigation to understand the current research progress, main achievements, existing problems, and future directions of MMOT. However, most existing MMOT reviews primarily focus on two modalities (\eg RGB+depth~\cite{tang2022survey,yang2022rgbd,zhou2024rgbd}, and RGB+thermal infrade~\cite{zhang2023review,tang2024revisiting}), and a comprehensive and in-depth investigation of object tracking involving more than two modalities is notably absent. We also note that a review focused on depth and thermal infrared modalities~\cite{zhang2024multi}, but it still does not cover the current popular MMOT tasks, \eg RGBL tracking and RGBE tracking. To fill this gap, we take the first step and conduct the first and most comprehensive investigation to date, covering various MMOT tasks\footnote{The various tasks we discuss in this paper are in single object tracking.} and providing researchers with a thorough perspective on the latest advancements in this field.

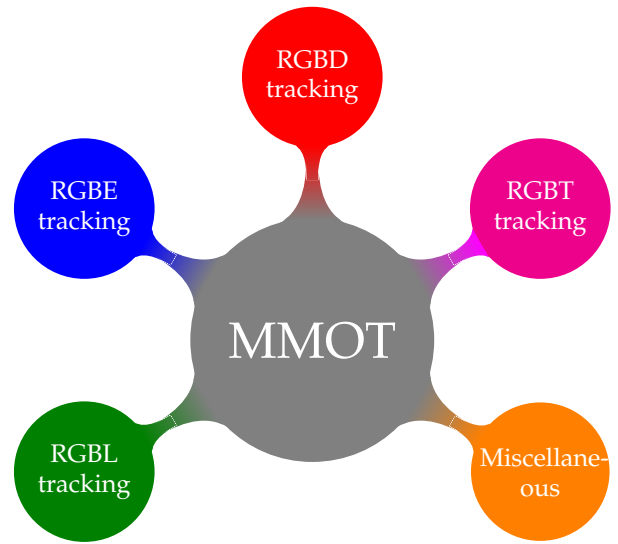
\begin{figure}[ht]
\centering
\tikzstyle{every node}=[font=\large, scale=0.8]
\begin{tikzpicture}[line width=1.0pt,scale=0.7]
  \path[mindmap,concept color=gray,text=white]
    node[font=\Huge, concept] {MMOT}
    [clockwise from=210]
    child[concept color=green!50!black] {
      node[concept] {RGBL tracking}
      [clockwise from=90]
    }  
    child[concept color=blue] {
      node[concept] {RGBE tracking}
      [clockwise from=-30]
    }
    child[concept color=red] { node[concept] {RGBD tracking} }
    child[concept color=magenta] { node[concept] {RGBT tracking} }
    child[concept color=orange] { node[concept] {Miscellane- ous}};
\end{tikzpicture}

    \caption{Scope of MMOT.}
    \label{fig:scope_of_MMOT}
\end{figure}

\begin{figure*}[t!]
  \centering

\includegraphics[width=1.0\linewidth]{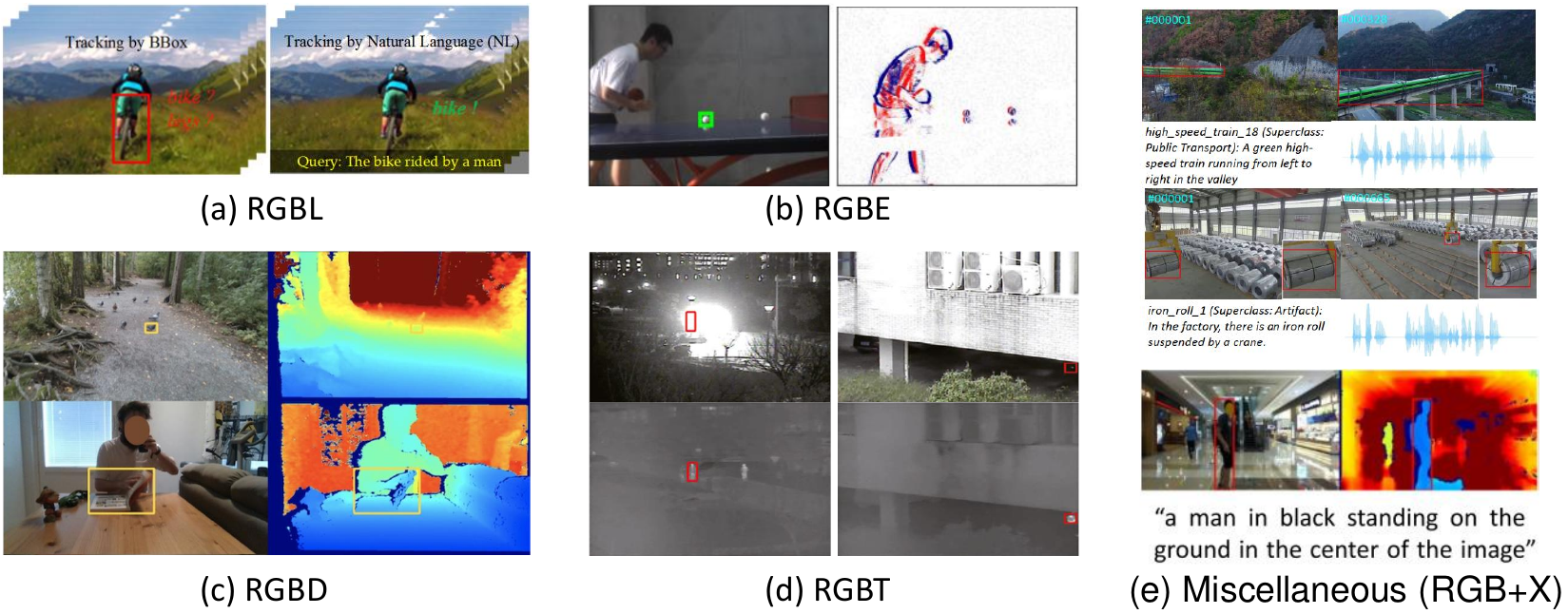}
  \caption{Data samples of five main MMOT tasks: (a) RGBL tracking, (b) RGBE tracking, (c) RGBD tracking, (d) RGBT tracking, and (e) miscellaneous (RGB+X) tracking. The figures are borrowed from~\cite{wang2021towards,zhu2024crsot,yan2021depthtrack,li2019rgb,zhang2022webuav,zhu2024unimod1k}, respectively. }
  \label{fig:examples}
\end{figure*}

\section{Scope of MMOT}
\label{sec:scope}

According to the different modalities used, we first divide existing MMOT tasks into 5 main categories: RGBL tracking, RGBE tracking, RGBD tracking, RGBT tracking, and miscellaneous (RGB+X), where X can be any modality, such as language, depth, and event. The taxonomy relations of different MMOT tasks are illustrated in Fig.~\ref{fig:scope_of_MMOT}. Some data samples of these MMOT tasks are shown in Fig.~\ref{fig:examples}.

We give the informal definitions of different MMOT tasks as follows: \textbf{1)} RGBL or Vision-language tracking is an advanced computer vision task that involves tracking objects in visual scenes based on the description in natural language, combining the capabilities of image recognition and natural language processing to understand and estimate the movement of objects across video sequences. \textbf{2)} RGBE tracking is a visual object tracking task that leverages the complementary information from both RGB (Red, Green, Blue) color images and event streams, which capture asynchronous events of motion changes, to enhance the tracking performance in environments with rapid motion or extreme lighting conditions. \textbf{3)} RGBD tracking is a visual object tracking technique that utilizes both RGB color information and depth (D) data to track objects in video sequences, providing enhanced accuracy and robustness, particularly in scenarios where depth information is crucial for understanding the scene. \textbf{4)} RGBT tracking is a multi-modal object tracking task that combines data from RGB color images and thermal (T) images to enhance the accuracy and robustness of tracking objects in various environments and lighting conditions. \textbf{5)} Miscellaneous (RGB+X) tracking refers to a class of multi-modal object tracking methods that can combine traditional RGB visual data with multiple additional 'X' modalities, such as thermal, depth, event, or language, to improve tracking performance and robustness across various environments and challenging conditions.

The widely used datasets and representative methods are summarized in Tabs.~\ref{tab:datasets} and~\ref{tab:paper_list}. Since MMOT is a rapidly evolving and promising field, we have launched this ``Awesome Multi-modal Object Tracking'' project on GitHub to keep track of the latest advancements in this area. All researchers are welcome to collaborate on this project. We hope this project can better promote the development of large multi-modal foundation tracking models and even artificial intelligence.

\bibliographystyle{IEEEtran}
\bibliography{main}

{
\scriptsize
\onecolumn
\begin{center}
    \begin{longtable}{|m{1.5cm}|m{1.5cm}|m{3.0cm}|m{3.0cm}|m{3.0cm}|m{3.0cm}|} 
	\caption{Summary of MMOT datasets.} \label{tab:datasets}  \\
         
	\hline 
        \multicolumn{1}{|c}{\textbf{Dataset}} &
        \multicolumn{1}{|c}{\textbf{Publish}} &
        \multicolumn{1}{|c}{\textbf{Title}} &
        \multicolumn{1}{|c}{\textbf{Project page}} &
        \multicolumn{1}{|c|}{\textbf{Code base}} &
        \multicolumn{1}{|c|}{\textbf{Introduction}}  \\ 
        \hline 

	\endfirsthead 

	\multicolumn{3}{c}%
	{{\bfseries \tablename\ \thetable{} -- continued from previous page}} \\
	\hline 
        \multicolumn{1}{|c}{\textbf{Dataset}} &
        \multicolumn{1}{|c}{\textbf{Publish}} &
        \multicolumn{1}{|c}{\textbf{Title}} &
        \multicolumn{1}{|c}{\textbf{Project page}} &
        \multicolumn{1}{|c|}{\textbf{Code base}} &
        \multicolumn{1}{|c|}{\textbf{Introduction}}  \\ 
        \hline 
 
	\endhead  
			
	\hline \multicolumn{3}{l}{{Continued on next page}} \\ 
	\endfoot  
			
	\hline
	\endlastfoot  

        \rowcolor[gray]{0.9} \multicolumn{6}{|c|}{\textbf{RGBL Tracking}} \\
        \hline
            
         OTB99-L &  CVPR-2017 & Tracking by Natural Language Specification & \url{https://github.com/QUVA-Lab/lang-tracker}  & \url{https://github.com/QUVA-Lab/lang-tracker}  & An early vision-language tracking dataset with 99 videos.   \\ 
       \hline
        
       LaSOT &  CVPR-2019 & LaSOT: A High-quality Benchmark for Large-scale Single Object Tracking & \url{http://vision.cs.stonybrook.edu/~lasot/}  & \url{https://github.com/HengLan/LaSOT_Evaluation_Toolkit}  & A large-scale dataset contains 1,400 video sequences with more than 3.5M frames.  \\   \hline

        LaSOT$\rm _{Ext}$  &  IJCV-2021 &  LaSOT: A High-quality Large-scale Single Object Tracking Benchmark & \url{http://vision.cs.stonybrook.edu/~lasot/}  & \url{https://github.com/HengLan/LaSOT_Evaluation_Toolkit}  & An expanded version of LaSOT, including 15 categories and 150 videos. \\   \hline

        TNL2K &  CVPR-2021 & WebUAV-3M: A Benchmark for Unveiling the Power of Million-Scale Deep UAV Tracking & \url{https://sites.google.com/view/langtrackbenchmark/}  & \url{https://github.com/wangxiao5791509/TNL2K_evaluation_toolkit}  & A large-scale dataset contains 2,000 videos and 1.24M frames.     \\ 
        \hline
        
        WebUAV-3M &  TPAMI-2023 & WebUAV-3M: A Benchmark for Unveiling the Power of Million-Scale Deep UAV Tracking & \url{https://github.com/983632847/WebUAV-3M}  & \url{https://github.com/983632847/WebUAV-3M}  & A large-scale multi-modal dataset for UAV tracking contains 3.3 million frames across 4,500 videos, with vision, language, and audio modalities.     \\   \hline

       MGIT & 	NeurIPS-2023  &   A Multi-modal Global Instance Tracking Benchmark (MGIT): Better Locating Target in Complex Spatio-temporal and Causal Relationship & \url{http://videocube.aitestunion.com/} & \url{https://github.com/huuuuusy/videocube-toolkit} & This dataset consists of 150 long video sequences, 2.03M frames, and three semantic grains (i.e., action, activity, and story).    \\ \hline

        VastTrack &  arXiv-2024 & VastTrack: Vast Category Visual Object Tracking & \url{https://github.com/HengLan/VastTrack}  & \url{https://github.com/HengLan/VastTrack}  & A dataset encompassing 2,115 categories, 50,510 videos, and totaling 4.2M frames. \\ \hline

        WebUOT-1M &  arXiv-2024 & WebUOT-1M: Advancing Deep Underwater Object Tracking with A Million-Scale Benchmark & \url{https://github.com/983632847/Awesome-Multimodal-Object-Tracking}  & \url{https://github.com/983632847/Awesome-Multimodal-Object-Tracking}  & The first million-scale underwater object tracking dataset contains 1,500 video sequences and 1.1 million frames. \\ \hline

        \rowcolor[gray]{0.9}  \multicolumn{6}{|c|}{\textbf{RGBE Tracking}} \\
        \hline
        
        FE108 & ICCV-2021  &   Object Tracking by Jointly Exploiting Frame and Event Domain & \url{https://zhangjiqing.com/dataset/} & \url{https://zhangjiqing.com/dataset/} & A dataset contains 108 videos, and 21 classes.  \\ \hline

        COESOT & arXiv-2022  &   Revisiting Color-Event based Tracking: A Unified Network, Dataset, and Metric & \url{https://github.com/Event-AHU/COESOT} & \url{https://github.com/Event-AHU/COESOT} & A large-scale RGBE dataset containing 1,354 RGB-event video pairs covering 90 target object categories. \\ \hline
        
        VisEvent & TC-2023  &   VisEvent: Reliable Object Tracking via Collaboration of Frame and Event Flows & \url{https://github.com/wangxiao5791509/VisEvent_SOT_Benchmark} & \url{https://github.com/wangxiao5791509/VisEvent_SOT_Benchmark} & A datasets contains 820 RGB-event video pairs.  \\ \hline

        EventVOT & CVPR-2023  &   Event Stream-based Visual Object Tracking: A High-Resolution Benchmark Dataset and A Novel Baseline  & \url{https://github.com/Event-AHU/EventVOT_Benchmark} & \url{https://github.com/Event-AHU/EventVOT_Benchmark} & The first high definition (1440x1080 and 1280x800) event-based dataset contains 1,141 event videos.  \\ \hline

        CRSOT & arXiv-2024  &   CRSOT: Cross-Resolution Object Tracking using Unaligned Frame and Event Cameras & \url{https://github.com/Event-AHU/Cross_Resolution_SOT} & \url{https://github.com/Event-AHU/Cross_Resolution_SOT} & A large-scale dataset for cross-resolution RGBE tracking with 1,030 RGB-event video pairs. \\ \hline

        FELT & arXiv-2024  &  Long-term Frame-Event Visual Tracking: Benchmark Dataset and Baseline  & \url{https://github.com/Event-AHU/FELT_SOT_Benchmark} & \url{https://github.com/Event-AHU/FELT_SOT_Benchmark} &  A long-term RGBE tracking dataset contains 742 RGB-event video pairs.  \\ \hline

        \rowcolor[gray]{0.9}  \multicolumn{6}{|c|}{\textbf{RGBD Tracking}} \\
        \hline
        
        PTB & ICCV-2013  &   Tracking Revisited using RGBD Camera: Unified Benchmark and Baselines & \url{https://tracking.cs.princeton.edu/index.html} & \url{https://tracking.cs.princeton.edu/eval.phpl} & A RGBD tracking dataset consists of 100 videos.  \\ \hline

        STC & TC-2018  &  Robust Fusion of Color and Depth Data for RGB-D Target Tracking Using Adaptive Range-Invariant Depth Models and Spatio-Temporal Consistency Constraints  &   \url{} & \url{https://beardatashare.bham.ac.uk/dl/fiVnhJRjkyNN8QjSAoiGSiBY/RGBDdataset.zip} & A dataset contains 36 video sequences.  \\ \hline

        CDTB & ICCV-2019  &   CDTB: A Color and Depth Visual Object Tracking Dataset and Benchmark  & \url{https://www.votchallenge.net/vot2019/dataset.html} & \url{https://www.votchallenge.net/vot2019/dataset.html} & A dataset contains 80 video sequences with more than 100,000 frames.  \\ \hline
        
        DepthTrack & ICCV-2021  &   DepthTrack: Unveiling the Power of RGBD Tracking & \url{https://github.com/xiaozai/DeT} & \url{https://github.com/xiaozai/DeT} &  This dataset contains 200 RGBD video sequences, 150 of which are used for training and 50 for testing. \\ \hline

        RGBD1K & AAAI-2023  &   RGBD1K: A Large-scale Dataset and Benchmark for RGB-D Object Tracking & \url{https://github.com/xuefeng-zhu5/RGBD1K} & \url{https://github.com/xuefeng-zhu5/RGBD1K} & A large-scale RGBD tracking dataset contains 1,050 video sequences.  \\ \hline

        DTTD & CVPRW-2023  &   Digital Twin Tracking Dataset (DTTD): A New RGB+Depth 3D Dataset for Longer-Range Object Tracking Applications & \url{https://github.com/augcog/DTTDv1} & \url{https://github.com/augcog/DTTDv1} & A RGBD tracking dataset contains 103 scenes of 10 common off-the-shelf objects.   \\ \hline

        ARKitTrack & CVPR-2023  &   3333 & \url{https://arkittrack.github.io/} & \url{https://github.com/lawrence-cj/ARKitTrack} & This dataset contains 300 RGBD video sequences, covering 455 objects, and the total number of frames reaches 229.7K.  \\ \hline

        \rowcolor[gray]{0.9}  \multicolumn{6}{|c|}{\textbf{RGBT Tracking}} \\
        \hline
        
        GTOT & TIP-2016  &   Learning Collaborative Sparse Representation for Grayscale-Thermal Tracking & \url{https://github.com/mmic-lcl/Datasets-and-benchmark-code}  & \url{https://pan.baidu.com/s/1QNidEo-HepRaS6OIZr7-Cw} & This dataset contains 50 grayscale and thermal infrared video pairs, covering 16 different scenes.  \\ \hline

        RGBT210 & ACM MM-2017  &   Weighted Sparse Representation Regularized Graph Learning for RGB-T Object Tracking & \url{https://github.com/mmic-lcl/Datasets-and-benchmark-code} & \url{https://drive.google.com/file/d/0B3i2rdXLNbdUTkhsLVRwcTBTMlU/view?resourcekey=0-vytg_w3hqlQfLhoiS2J8Dg} & This dataset contains 210 pairs of highly aligned RGBT video sequences, with a total of approximately 210K frames.  \\ \hline  

        RGBT234 & PR-2018  &   RGB-T Object Tracking:Benchmark and Baseline & \url{https://sites.google.com/view/ahutracking001/} & \url{https://sites.google.com/view/ahutracking001/} & This dataset is the extension of RGBT210 containing 234 video pairs.  \\ \hline
        
        LasHeR & TIP-2021  &   LasHeR: A Large-Scale High-Diversity Benchmark for RGBT Tracking & \url{https://github.com/BUGPLEASEOUT/LasHeR} & \url{https://github.com/BUGPLEASEOUT/LasHeR} & This dataset contains 1224 pairs of RGB visible and thermal infrared video sequences, with a total number of frames over 730K  \\ \hline

        VTUAV & CVPR-2022	  &   Visible-Thermal UAV Tracking: A Large-Scale Benchmark and New Baseline & \url{https://zhang-pengyu.github.io/DUT-VTUAV/} & \url{https://github.com/zhang-pengyu/DUT-VTUAV} & This is a large-scale visible-thermal infrared multi-modal drone tracking dataset, containing 500 video sequences with a total of 1,664,549 frames of visible and thermal infrared image pairs at a resolution of 1920x1080.  \\ \hline

        MV-RGBT & arXiv-2024	  &   Revisiting RGBT Tracking Benchmarks from the Perspective of Modality Validity: A New Benchmark, Problem, and Method & \url{https://github.com/Zhangyong-Tang/MoETrack} & \url{https://github.com/Zhangyong-Tang/MoETrack} & This dataset covers 122 video pairs with a total of 89.9k frame pairs at a resolution of 640x480.  \\ \hline

        \rowcolor[gray]{0.9}  \multicolumn{6}{|c|}{\textbf{Miscellaneous (RGB+X) Tracking}} \\
        \hline
        
        WebUAV-3M &  TPAMI-2023 & WebUAV-3M: A Benchmark for Unveiling the Power of Million-Scale Deep UAV Tracking & \url{https://github.com/983632847/WebUAV-3M}  & \url{https://github.com/983632847/WebUAV-3M}  & A large-scale multi-modal dataset for UAV tracking contains 3.3 million frames across 4,500 videos, with vision, language, and audio modalities.     \\   \hline

        UniMod1K & IJCV-2024  &   UniMod1K: Towards a More Universal Large-Scale Dataset and Benchmark for Multi-modal Learning & \url{https://github.com/xuefeng-zhu5/UniMod1K} & \url{https://github.com/xuefeng-zhu5/UniMod1K} & This dataset contains 1050 video pairs, 2.5 million frames, with vision, depth and language modalities.  \\ \hline

    \end{longtable}
\end{center}
}

{
\scriptsize
\onecolumn
\begin{center}
    \begin{longtable}{|m{1.5cm}|m{1.5cm}|m{4.0cm}|m{4.0cm}|m{4.0cm}|} 
	\caption{Paper list for MMOT.} \label{tab:paper_list}  \\
         
	\hline 
         \multicolumn{1}{|c}{\textbf{Method}} &
        \multicolumn{1}{|c}{\textbf{Publish}} &
        \multicolumn{1}{|c}{\textbf{Title}} &
        \multicolumn{1}{|c}{\textbf{Paper link}} &
        \multicolumn{1}{|c|}{\textbf{Code base}} \\ 
        \hline 
    
	\endfirsthead 

	\multicolumn{3}{c}%
	{{\bfseries \tablename\ \thetable{} -- continued from previous page}} \\
	\hline 
        \multicolumn{1}{|c}{\textbf{Method}} &
        \multicolumn{1}{|c}{\textbf{Publish}} &
        \multicolumn{1}{|c}{\textbf{Title}} &
        \multicolumn{1}{|c}{\textbf{Project page}} &
        \multicolumn{1}{|c|}{\textbf{Code base}} \\ 
        \hline 
 
	\endhead  
			
	\hline \multicolumn{3}{l}{{Continued on next page}} \\ 
	\endfoot  
			
	\hline
	\endlastfoot  

       \rowcolor[gray]{0.9}  \multicolumn{5}{|c|}{{\textbf{RGBL Tracking}}} \\
        \hline

        DTLLM-VLT &  CVPRW-2024 & DTLLM-VLT: Diverse Text Generation for Visual Language Tracking Based on LLM
 &  \url{https://arxiv.org/abs/2405.12139}  & \url{}       \\ \hline

        UVLTrack &  AAAI-2024 & Unifying Visual and Vision-Language Tracking via Contrastive Learning &  \url{https://arxiv.org/abs/2401.11228}  & \url{https://github.com/OpenSpaceAI/UVLTrack}       \\ \hline
        
        QueryNLT &  CVPR-2024 & Context-Aware Integration of Language and Visual References for Natural Language Tracking &  \url{https://arxiv.org/abs/2403.19975}  & \url{https://github.com/twotwo2/QueryNLT}       \\ 
        \hline
        
        OSDT &  TCSVT-2024  & One-Stream Stepwise Decreasing for Vision-Language Tracking &  \url{https://ieeexplore.ieee.org/abstract/document/10510485}  & \url{}       \\ 
        \hline

        TTCTrack &  ICASSP-2024  & Textual Tokens Classification for Multi-Modal Alignment in Vision-Language Tracking &  \url{https://ieeexplore.ieee.org/document/10446122}  & \url{}       \\ 
        \hline
              
        MMTrack &  TCSVT-2024  & Toward Unified Token Learning for Vision-Language Tracking &  \url{https://ieeexplore.ieee.org/abstract/document/10208210}  & \url{}       \\ 
        \hline
        
        Ye \etal &  Remote Sensing-2024  & Multimodal Features Alignment for Vision–Language Object Tracking &  \url{https://www.mdpi.com/2072-4292/16/7/1168}  & \url{}       \\ 
        \hline

        All in One &  ACM MM-2023  & All in One: Exploring Unified Vision-Language Tracking with Multi-Modal Alignment &  \url{https://arxiv.org/abs/2307.03373}  & \url{https://github.com/983632847/All-in-One}       \\ 
        \hline
              
        CiteTracker &  ICCV-2023  & CiteTracker: Correlating Image and Text for Visual Tracking &  \url{https://arxiv.org/abs/2308.11322}  & \url{https://github.com/NorahGreen/CiteTracker}       \\ 
        \hline
        
        JointNLT &  CVPR-2023  & Joint Visual Grounding and Tracking with Natural Language Specification &  \url{https://arxiv.org/abs/2303.12027#:~:text=Tracking%20by%20natural%20language%20specification%20aims%20to%20locate,tracking%20model%20to%20implement%20these%20two%20steps%2C%20respectively.}  & \url{https://github.com/lizhou-cs/JointNLT}       \\ 
        \hline
        
        DecoupleTNL &   ICCV-2023  & Tracking by Natural Language Specification with Long Short-term Context Decoupling &  \url{https://ieeexplore.ieee.org/document/10378598/references#references}  & \url{}       \\ 
        \hline

        Zhao \etal &  PRL-2023  & Transformer vision-language tracking via proxy token guided cross-modal fusion &  \url{https://www.sciencedirect.com/science/article/abs/pii/S0167865523000545}  & \url{}       \\ 
        \hline
              
        OVLM &  TMM-2023  & One-Stream Vision-Language Memory Network for Object Tracking &  \url{https://ieeexplore.ieee.org/document/10149530}  & \url{}       \\ 
        \hline
        
        SATracker &  ArXiv-2023  & Target-Centric Semantics for Vision-Language Tracking &  \url{https://arxiv.org/abs/2311.17085}  & \url{}       \\ 
        \hline
        
        VLATrack &  RICAI-2023  & Multi-Modal Object Tracking with Vision-Language Adaptive Fusion and Alignment &  \url{https://ieeexplore.ieee.org/document/10489325}  & \url{}       \\ 
        \hline

        VLT$\rm _{TT}$ & ArXiv-2023 & Divert More Attention to Vision-Language Object Tracking &  \url{https://arxiv.org/abs/2307.10046}  & \url{https://github.com/JudasDie/SOTS}       \\ 
        \hline

        VLT$\rm _{TT}$ &  NeurIPS-2022 & Divert More Attention to Vision-Language Tracking &  \url{https://arxiv.org/abs/2207.01076}  & \url{https://github.com/JudasDie/SOTS}       \\ 
        \hline
              
        AdaRS &  CVPRW-22 & Cross-modal Target Retrieval for Tracking by Natural Language &  \url{https://ieeexplore.ieee.org/document/9857151}  & \url{}       \\ 
        \hline
        
        SNLT &  CVPR-2021  & Siamese Natural Language Tracker: Tracking by Natural Language Descriptions with Siamese Trackers &  \url{https://arxiv.org/abs/1912.02048}  & \url{https://github.com/fredfung007/snlt}       \\ 
        \hline

        \rowcolor[gray]{0.9}  \multicolumn{5}{|c|}{{\textbf{RGBE Tracking}}} \\
        \hline
        
        Mamba-FETrack &  ArXiv-2024 & Mamba-FETrack: Frame-Event Tracking via State Space Model &  \url{https://arxiv.org/abs/2404.18174}  & \url{https://github.com/Event-AHU/Mamba_FETrack}       \\ 
        \hline

        AMTTrack &  ArXiv-2024 & Long-term Frame-Event Visual Tracking: Benchmark Dataset and Baseline &  \url{https://arxiv.org/abs/2401.02826}  & \url{https://github.com/Event-AHU/FELT_SOT_Benchmark}       \\ 
        \hline
              
        TENet &  ArXiv-2024 & TENet: Targetness Entanglement Incorporating with Multi-Scale Pooling and Mutually-Guided Fusion for RGB-E Object Tracking &  \url{https://arxiv.org/abs/2405.05004}  & \url{https://github.com/SSSpc333/TENet}       \\ 
        \hline
        
        HDETrack &  CVPR-2024  & Event Stream-based Visual Object Tracking: A High-Resolution Benchmark Dataset and A Novel Baseline &  \url{https://arxiv.org/abs/2309.14611}  & \url{https://github.com/Event-AHU/EventVOT_Benchmark}       \\ 
        \hline
        
        Zhu \etal &  ArXiv-2024 & CRSOT: Cross-Resolution Object Tracking using Unaligned Frame and Event Cameras &  \url{https://arxiv.org/pdf/2403.05839.pdf}  & \url{https://github.com/Event-AHU/FELT_SOT_Benchmark}       \\ 
        \hline

        CDFI &  ArXiv-2024 & Object Tracking by Jointly Exploiting Frame and Event Domain &  \url{https://arxiv.org/abs/2109.09052}  & \url{}       \\ 
        \hline

        MMHT &  ArXiv-2024 & Reliable Object Tracking by Multimodal Hybrid Feature Extraction and Transformer-Based Fusion &  \url{https://arxiv.org/abs/2405.17903}  & \url{}       \\ 
        \hline
              
        Zhu \etal  &  ICCV-2023  & Cross-modal Orthogonal High-rank Augmentation for RGB-Event Transformer-trackers &  \url{https://arxiv.org/abs/2307.04129}  & \url{https://github.com/ZHU-Zhiyu/High-Rank_RGB-Event_Tracker}       \\ 
        \hline
        
        AFNet &  CVPR-2023 & Frame-Event Alignment and Fusion Network for High Frame Rate Tracking &  \url{https://arxiv.org/abs/2305.15688}  & \url{https://github.com/Jee-King/AFNet}       \\ 
        \hline
        
        RT-MDNet &  TC-2023  & VisEvent: Reliable Object Tracking via Collaboration of Frame and Event Flows &  \url{https://arxiv.org/abs/2108.05015}  & \url{https://github.com/wangxiao5791509/VisEvent_SOT_Benchmark}       \\ 
        \hline

        Event-tracking &  NeurIPS-2022  & Learning Graph-embedded Key-event Back-tracing for Object Tracking in Event Clouds &  \url{https://dl.acm.org/doi/10.5555/3600270.3600812}  & \url{https://github.com/ZHU-Zhiyu/Event-tracking}       \\ 
        \hline
              
        STNet &  CVPR-2022  & Spiking Transformers for Event-based Single Object Tracking &  \url{https://ieeexplore.ieee.org/document/9879994}  & \url{https://github.com/Jee-King/CVPR2022_STNet}       \\ 
        \hline
        
        CEUTrack &  ArXiv-2022 & Revisiting Color-Event based Tracking: A Unified Network, Dataset, and Metric &  \url{https://arxiv.org/abs/2211.11010}  & \url{https://github.com/Event-AHU/COESOT}       \\ 
        \hline
        
        CFE &  The Visual Computer-2021  & Multi-domain Collaborative Feature Representation for Robust Visual Object Tracking &  \url{https://arxiv.org/abs/2108.04521}  & \url{}       \\ 
        \hline

        \rowcolor[gray]{0.9}  \multicolumn{5}{|c|}{{\textbf{RGBD Tracking}}} \\
        \hline

        SSLTrack &  PR-2024  & Self-supervised learning for RGB-D object tracking &  \url{https://www.sciencedirect.com/science/article/pii/S0031320324002942}  & \url{}       \\ 
        \hline
              
        VADT &  ICASSP-2024  & Visual Adapt for RGBD Tracking &  \url{https://ieeexplore.ieee.org/document/10447728}  & \url{}       \\ 
        \hline
        
        FECD &  PRL-2024  & Feature enhancement and coarse-to-fine detection for RGB-D tracking &  \url{https://www.sciencedirect.com/science/article/pii/S0167865524000412}  & \url{}       \\ 
        \hline
        
        CDAAT &  SPL-2024  & Adaptive Colour-Depth Aware Attention for RGB-D Object Tracking &  \url{https://ieeexplore.ieee.org/document/10472092/}  & \url{https://github.com/xuefeng-zhu5/CDAAT}       \\ 
        \hline

        SPT &  AAAI-2023  & RGBD1K: A Large-scale Dataset and Benchmark for RGB-D Object Tracking &  \url{https://arxiv.org/pdf/2208.09787.pdf}  & \url{https://github.com/xuefeng-zhu5/RGBD1K}       \\ 
        \hline
              
        EMT &  CVPR-2023  & Resource-Efficient RGBD Aerial Tracking &  \url{https://ieeexplore.ieee.org/document/10204937/}  & \url{https://github.com/yjybuaa/RGBDAerialTracking}       \\ 
        \hline
        
        Track-it-in-3D &  ECCV-2022  & Towards Generic 3D Tracking in RGBD Videos: Benchmark and Baseline &  \url{https://link.springer.com/chapter/10.1007/978-3-031-20047-2_7}  & \url{https://github.com/yjybuaa/Track-it-in-3D}       \\ 
        \hline
        
        DMTracker &   ECCVW-2022  & Learning Dual-Fused Modality-Aware Representations for RGBD Tracking &  \url{https://arxiv.org/abs/2211.03055}  & \url{}       \\ 
        \hline

        DeT &  ICCV-2021  & DepthTrack: Unveiling the Power of RGBD Tracking &  \url{https://arxiv.org/abs/2108.13962}  & \url{https://github.com/xiaozai/DeT}       \\ 
        \hline
              
        TSDM &  ICPR-2021  & TSDM: Tracking by SiamRPN++ with a Depth-refiner and a Mask-generator &  \url{https://arxiv.org/ftp/arxiv/papers/2005/2005.04063.pdf}  & \url{https://github.com/lql-team/TSDM}       \\ 
        \hline
        
        3s-RGBD &  Neurocomputing-2021  & Single-scale siamese network based RGB-D object tracking with adaptive bounding boxes &  \url{https://www.sciencedirect.com/sdfe/reader/pii/S0925231221005439/pdf}  & \url{}       \\ 
        \hline
        
        DAL &  ICPR-2020  & DAL : A deep depth-aware long-term tracker &  \url{https://arxiv.org/abs/1912.00660}  & \url{https://github.com/xiaozai/DAL}       \\ 
        \hline

        RF-CFF &  Applied Soft Computing Journal-2020  & Robust fusion for RGB-D tracking using CNN features &  \url{https://www.sciencedirect.com/sdfe/reader/pii/S1568494620302428/pdf}  & \url{}       \\ 
        \hline
              
        SiamOC &  ICSP-2020  & An Occlusion-Aware RGB-D Visual Object Tracking Method Based on Siamese Network &  \url{https://ieeexplore.ieee.org/stamp/stamp.jsp?tp=&arnumber=9320907}  & \url{}       \\ 
        \hline
        
        WCO &  Sensors-2020  & Robust RGBD Tracking via Weighted Convolution Operators &  \url{https://ieeexplore.ieee.org/stamp/stamp.jsp?tp=&arnumber=8950173/}  & \url{}       \\ 
        \hline
        
        OTR &  CVPR-2019  & Object Tracking by Reconstruction with View-Specific Discriminative Correlation Filters &  \url{https://openaccess.thecvf.com/content_CVPR_2019/papers/Kart_Object_Tracking_by_Reconstruction_With_View-Specific_Discriminative_Correlation_Filters_CVPR_2019_paper.pdf}  & \url{https://github.com/ugurkart/OTR}       \\ 
        \hline

        H-FCN &  Information Fusion-2019  & Hierarchical multi-modal fusion FCN with attention model for RGB-D tracking &  \url{https://www.sciencedirect.com/sdfe/reader/pii/S1566253517306784/pdf}  & \url{}       \\ 
        \hline
              
        Kuai \etal &  IEEE Sensors Journal-2019  & Target-Aware Correlation Filter Tracking in RGBD Videos &  \url{https://ieeexplore.ieee.org/abstract/document/8752050}  & \url{}       \\ 
        \hline
        
        RGBD-OD &  CIS-2019  & RGB-D Object Tracking with Occlusion Detection &  \url{https://ieeexplore.ieee.org/stamp/stamp.jsp?tp=&arnumber=9023755}  & \url{}       \\ 
        \hline
        
        3DMS &  ICST-2019  & Exploiting Depth Information to Increase Object Tracking Robustness &  \url{https://ieeexplore.ieee.org/stamp/stamp.jsp?tp=&arnumber=8861628/}  & \url{}       \\ 
        \hline

        CA3DMS &  TMM-2019  & Context-Aware Three-Dimensional Mean-Shift With Occlusion Handling for Robust Object Tracking in RGB-D Videos &  \url{https://ieeexplore.ieee.org/stamp/stamp.jsp?tp=&arnumber=8425768}  & \url{https://github.com/yeliu2013/ca3dms-toh}       \\ 
        \hline
              
        Depth-CCF &  GSKI-2019  & Depth Information Aided Constrained correlation Filter for Visual Tracking &  \url{https://iopscience.iop.org/article/10.1088/1755-1315/234/1/012005}  & \url{}       \\ 
        \hline
        
        STC &  TC-2018  & Robust Fusion of Color and Depth Data for RGB-D Target Tracking Using Adaptive Range-Invariant Depth Models and Spatio-Temporal Consistency Constraints &  \url{https://ieeexplore.ieee.org/stamp/stamp.jsp?tp=&arnumber=8026575}  & \url{https://github.com/shine636363/RGBDtracker}       \\ 
        \hline
        
        Kart \etal &  ECCVW-2018  & How to Make an RGBD Tracker? &  \url{https://link.springer.com/chapter/10.1007/978-3-030-11009-3_8}  & \url{https://github.com/ugurkart/rgbdconverter}       \\ 
        \hline

        Leng \etal &  IEEE Access-2018  & Real-Time RGB-D Visual Tracking With Scale Estimation and Occlusion Handling &  \url{https://ieeexplore.ieee.org/document/8353501}  & \url{}       \\ 
        \hline
              
        DM-DCF &  ICPR-2018   & Depth Masked Discriminative Correlation Filter &  \url{https://arxiv.org/pdf/1802.09227.pdf}  & \url{}       \\ 
        \hline
        
        OACPF &  Access-2018  & Occlusion-Aware Correlation Particle FilterTarget Tracking Based on RGBD Data &  \url{https://ieeexplore.ieee.org/stamp/stamp.jsp?tp=&arnumber=8463446}  & \url{}       \\ 
        \hline
        
        RT-KCF &  CCDC-2018  & A Real-time RGB-D tracker based on KCF &  \url{https://ieeexplore.ieee.org/stamp/stamp.jsp?tp=&arnumber=8407972}  & \url{}       \\ 
        \hline

        ODIOT &  Neural Process Letters-2017  & Online Depth Image-Based Object Tracking with Sparse Representation and Object Detection &  \url{https://link.springer.com/content/pdf/10.1007/s11063-016-9509-y.pdf}  & \url{}       \\ 
        \hline
              
        ROTSL &  ITEE-2017  & Robust Object Tracking with RGBD-based Sparse Learning &  \url{https://link.springer.com/article/10.1631/FITEE.1601338}  & \url{}       \\ 
        \hline
        
        DLS &  ICPR-2016  & Online RGB-D Tracking via Detection-Learning-Segmentation &  \url{https://ieeexplore.ieee.org/stamp/stamp.jsp?tp=&arnumber=7899805}  & \url{}       \\ 
        \hline
        
        DS-KCF\_shape &  RTIP-2016  & DS-KCF: A Real-time Tracker for RGB-D Data &  \url{https://link.springer.com/content/pdf/10.1007/s11554-016-0654-3.pdf}  & \url{https://github.com/mcamplan/DSKCF_JRTIP2016}       \\ 
        \hline

        3D-T &  CVPR-2016  & 3D Part-Based Sparse Tracker with Automatic Synchronization and Registration &  \url{https://www.cv-foundation.org/openaccess/content_cvpr_2016/papers/Bibi_3D_Part-Based_Sparse_CVPR_2016_paper.pdf}  & \url{https://github.com/adelbibi/3D-Part-Based-Sparse-Tracker-with-Automatic-Synchronization-and-Registration}       \\ 
        \hline
              
        OAPF &  CVIU-2016  & Occlusion Aware Particle Filter Tracker to Handle Complex and Persistent Occlusions &  \url{http://ishiilab.jp/member/meshgi-k/files/ai/prl14/OAPF.pdf}  & \url{}       \\ 
        \hline
        
        CDG &  CAC-2015  & Using Consistency of Depth Gradient to Improve Visual Tracking in RGB-D sequences &  \url{https://ieeexplore.ieee.org/document/7382555}  & \url{}       \\ 
        \hline
        
        DS-KCF &  BMVC-2015  & Real-time RGB-D Tracking with Depth Scaling Kernelised Correlation Filters and Occlusion Handling &  \url{https://core.ac.uk/reader/78861956}  & \url{https://github.com/mcamplan/DSKCF_BMVC2015}       \\ 
        \hline

        DOHR &  FSKD-2015  & Robust Object Tracking Using Color and Depth Images with a Depth Based Occlusion Handling and Recovery &  \url{https://ieeexplore.ieee.org/document/7382068}  & \url{}       \\ 
        \hline
              
        ISOD &  SP-2015  & 3D Object Tracking via Image Sets and Depth-Based Occlusion Detection &  \url{https://www.sciencedirect.com/science/article/pii/S0165168414004204}  & \url{}       \\ 
        \hline
        
        OL3DC &  Neurocomputing-2015   & Online Learning 3D Context for Robust Visual Tracking &  \url{https://www.sciencedirect.com/science/article/pii/S0925231214013757}  & \url{}       \\ 
        \hline
        
        MCBT &  Neurocomputing-2014  & Multi-Cue Based Tracking &  \url{http://citeseerx.ist.psu.edu/viewdoc/download?doi=10.1.1.700.8771&rep=rep1&type=pdf}  & \url{}       \\ 
        \hline

        PT &  ICCV-2013  & Tracking Revisited using RGBD Camera: Unified Benchmark and Baselines &  \url{https://vision.princeton.edu/projects/2013/tracking/paper.pdf}  & \url{https://tracking.cs.princeton.edu/index.html}       \\ 
        \hline
              
        Matteo \etal &  IROS-2012  & Tracking people within groups with RGB-D data &  \url{https://ieeexplore.ieee.org/abstract/document/6385772/}  & \url{}       \\ 
        \hline
        
        AMCT &  JDOS-2012  & Adaptive Multi-cue 3D Tracking of Arbitrary Objects &  \url{https://link.springer.com/chapter/10.1007/978-3-642-32717-9_36}  & \url{}       \\ 
        \hline

        \rowcolor[gray]{0.9}  \multicolumn{5}{|c|}{{\textbf{RGBT Tracking}}} \\
        \hline
        
        GMMT &  AAAI-2024  & Generative-based Fusion Mechanism for Multi-Modal Tracking &  \url{https://arxiv.org/abs/2309.01728}  & \url{https://github.com/Zhangyong-Tang/GMMT}       \\ 
        \hline

        BAT &  AAAI-2024  & Bi-directional Adapter for Multi-modal Tracking &  \url{https://arxiv.org/abs/2312.10611}  & \url{https://github.com/SparkTempest/BAT}       \\ 
        \hline
              
        ProFormer &  TCSVT-2024  & RGBT Tracking via Progressive Fusion Transformer with Dynamically Guided Learning &  \url{https://ieeexplore.ieee.org/document/10506555/}  & \url{}       \\ 
        \hline
        
        QueryTrack &  TIP-2024  & QueryTrack: Joint-Modality Query Fusion Network for RGBT Tracking &  \url{https://ieeexplore.ieee.org/document/10516307}  & \url{}       \\ 
        \hline
        
        CAT++ &  TIP-2024  & RGBT Tracking via Challenge-Based Appearance Disentanglement and Interaction &  \url{https://ieeexplore.ieee.org/abstract/document/10460420}  & \url{}       \\ 
        \hline

        TATrack &  ArXiv-2024 & Temporal Adaptive RGBT Tracking with Modality Prompt &  \url{https://arxiv.org/abs/2401.01244}  & \url{}       \\ 
        \hline
              
        MArMOT &  ArXiv-2024 & Cross-Modal Object Tracking: Modality-Aware Representations and A Unified Benchmark &  \url{https://arxiv.org/abs/2111.04264}  & \url{}       \\ 
        \hline
        
        AMNet &  TCSVT-2024  & AMNet: Learning to Align Multi-modality for RGB-T Tracking &  \url{https://ieeexplore.ieee.org/abstract/document/10472533}  & \url{}       \\ 
        \hline
        
        MCTrack &  TCSVT-2024  & Towards Modalities Correlation for RGB-T Tracking &  \url{https://ieeexplore.ieee.org/abstract/document/10517645}  & \url{}       \\ 
        \hline

        AFter &  ArXiv-2024 & AFter: Attention-based Fusion Router for RGBT Tracking &  \url{https://arxiv.org/abs/2405.02717}  & \url{https://github.com/Alexadlu/AFter}       \\ 
        \hline
              
        CSTNet &  ArXiv-2024 & Transformer-based RGB-T Tracking with Channel and Spatial Feature Fusion &  \url{https://arxiv.org/abs/2405.03177}  & \url{https://github.com/LiYunfengLYF/CSTNet}       \\ 
        \hline
        
        TBSI &  CVPR-2023  & Bridging Search Region Interaction with Template for RGB-T Tracking &  \url{https://openaccess.thecvf.com/content/CVPR2023/papers/Hui_Bridging_Search_Region_Interaction_With_Template_for_RGB-T_Tracking_CVPR_2023_paper.pdf}  & \url{https://github.com/RyanHTR/TBSI}       \\ 
        \hline
        
        DFNet &  TITS-2023  & Dynamic Fusion Network for RGBT Tracking &  \url{https://arxiv.org/abs/2109.07662}  & \url{https://github.com/PengJingchao/DFNet}       \\ 
        \hline

        CMD &  CVPR-2023  & Efficient RGB-T Tracking via Cross-Modality Distillation &  \url{https://ieeexplore.ieee.org/document/10205202}  & \url{}       \\ 
        \hline
              
        DFAT &  Information Fusion-2023  & Exploring fusion strategies for accurate RGBT visual object tracking &  \url{https://arxiv.org/abs/2201.08673}  & \url{https://github.com/Zhangyong-Tang/DFAT}       \\ 
        \hline
        
        QAT &  ACM MM-2023  & Quality-Aware RGBT Tracking via Supervised Reliability Learning and Weighted Residual Guidance &  \url{https://dl.acm.org/doi/10.1145/3581783.3612341}  & \url{}       \\ 
        \hline
        
        GuideFuse &  TIM-2023  & GuideFuse: A Novel Guided Auto-Encoder Fusion Network for Infrared and Visible Images &  \url{https://ieeexplore.ieee.org/document/10330731}  & \url{}       \\ 
        \hline
              
        MPLT &  ArXiv-2023 & RGB-T Tracking via Multi-Modal Mutual Prompt Learning &  \url{https://arxiv.org/abs/2308.16386}  & \url{https://github.com/HusterYoung/MPLT}       \\ 
        \hline
        
        HMFT &  CVPR-2022   & Visible-Thermal UAV Tracking: A Large-Scale Benchmark and New Baseline &  \url{https://arxiv.org/abs/2204.04120}  & \url{https://github.com/zhang-pengyu/HMFT}       \\ 
        \hline
        
        MFGNet &  TMM-2022  & MFGNet: Dynamic Modality-Aware Filter Generation for RGB-T Tracking &  \url{https://arxiv.org/abs/2107.10433}  & \url{https://github.com/wangxiao5791509/MFG_RGBT_Tracking_PyTorch}       \\ 
        \hline

        MBAFNet &  IEEE Sensors Journal-2022   & Multibranch Adaptive Fusion Network for RGBT Tracking  &  \url{https://ieeexplore.ieee.org/document/9721310}  & \url{}       \\ 
        \hline
              
        AGMINet &  TIM-2022   & Asymmetric Global–Local Mutual Integration Network for RGBT Tracking &  \url{https://ieeexplore.ieee.org/abstract/document/9840392/}  & \url{}       \\ 
        \hline
        
        APFNet &  AAAI-2022  & Attribute-Based Progressive Fusion Network for RGBT Tracking &  \url{https://cdn.aaai.org/ojs/20187/20187-13-24200-1-2-20220628.pdf}  & \url{https://github.com/yangmengmeng1997/APFNet}       \\ 
        \hline
        
        DMCNet &  TNNLS-2022  & Duality-Gated Mutual Condition Network for RGBT Tracking &  \url{https://ieeexplore.ieee.org/document/9737634}  & \url{}       \\ 
        \hline

        TFNet &  TCSVT-2022  & RGBT Tracking by Trident Fusion Network &  \url{https://ieeexplore.ieee.org/document/9383014}  & \url{}       \\ 
        \hline
              
        Feng \etal &  KBS-2022   & Learning reliable modal weight with transformer for robust RGBT tracking &  \url{https://www.sciencedirect.com/science/article/pii/S0950705122004579}  & \url{}       \\ 
        \hline
        
        JMMAC &  TIP-2021   & Jointly Modeling Motion and Appearance Cues for Robust RGB-T Tracking &  \url{https://ieeexplore.ieee.org/document/9364880/}  & \url{https://github.com/zhang-pengyu/JMMAC}       \\ 
        \hline
        
        ADRNet &  IJCV-2021  & Learning Adaptive Attribute-Driven Representation for Real-Time RGB-T Tracking &  \url{https://github.com/zhang-pengyu/ADRNet/blob/main/Zhang_IJCV2021_ADRNet.pdf}  & \url{https://github.com/zhang-pengyu/ADRNet}       \\ 
        \hline

        SiamCDA &  TCSVT-2021  & SiamCDA: Complementarity-and distractor-aware RGB-T tracking based on Siamese network &  \url{https://ieeexplore.ieee.org/abstract/document/9399460/}  & \url{https://github.com/Tianlu-Zhang/LSS-Dataset}       \\ 
        \hline
              
        Wang \etal &  TITS-2021   & Adaptive Fusion CNN Features for RGBT Object Tracking &  \url{https://ieeexplore.ieee.org/document/9426573}  & \url{}       \\ 
        \hline
        
        M$^5$L &  TIP-2021  & M$^5$L: Multi-Modal Multi-Margin Metric Learning for RGBT Tracking &  \url{https://arxiv.org/abs/2003.07650}  & \url{}       \\ 
        \hline
        
        CBPNet &  TMM-2021   & Multimodal Cross-Layer Bilinear Pooling for RGBT Tracking &  \url{https://ieeexplore.ieee.org/document/9340007/}  & \url{}       \\ 
        \hline

        MANet++ &  TIP-2021  & RGBT Tracking via Multi-Adapter Network with Hierarchical Divergence Loss &  \url{https://arxiv.org/abs/2011.07189}  & \url{}       \\ 
        \hline
              
        CMR &  TNNLS-2021   & RGBT Tracking via Noise-Robust Cross-Modal Ranking &  \url{https://ieeexplore.ieee.org/document/9406193/}  & \url{}       \\ 
        \hline
        
        GCMP &  Neurocomputing-2021   & RGBT tracking via cross-modality message passing  &  \url{https://dl.acm.org/doi/10.1016/j.neucom.2021.08.012}  & \url{}       \\ 
        \hline
        
        HDINet &  IEEE Sensors Journal-2021   & HDINet: Hierarchical Dual-Sensor Interaction Network for RGBT Tracking  &  \url{https://ieeexplore.ieee.org/abstract/document/9426927}  & \url{}       \\ 
        \hline

        CMPP &  CVPR-2020   & Cross-Modal Pattern-Propagation for RGB-T Tracking &  \url{https://openaccess.thecvf.com/content_CVPR_2020/papers/Wang_Cross-Modal_Pattern-Propagation_for_RGB-T_Tracking_CVPR_2020_paper.pdf}  & \url{}       \\ 
        \hline
              
        CAT &  ECCV-2020  & Challenge-Aware RGBT Tracking &  \url{https://ar5iv.labs.arxiv.org/abs/2007.13143}  & \url{}       \\ 
        \hline
        
        FANet &  TIV-2020   & FANet: Quality-Aware Feature Aggregation Network for Robust RGB-T Tracking &  \url{https://arxiv.org/abs/1811.09855}  & \url{}       \\ 
        \hline
        
        mfDiMP &   ICCVW-2019  & Multi-Modal Fusion for End-to-End RGB-T Tracking &  \url{https://arxiv.org/abs/1908.11714}  & \url{https://github.com/zhanglichao/end2end_rgbt_tracking}       \\ 
        \hline

        DAPNet &  ACM MM-2019  & Dense Feature Aggregation and Pruning for RGBT Tracking &  \url{https://arxiv.org/abs/1907.10451}  & \url{}       \\ 
        \hline
              
        DAFNet &  ICCVW-2019   & Deep Adaptive Fusion Network for High Performance RGBT Tracking &  \url{https://openaccess.thecvf.com/content_ICCVW_2019/html/VISDrone/Gao_Deep_Adaptive_Fusion_Network_for_High_Performance_RGBT_Tracking_ICCVW_2019_paper.html}  & \url{https://github.com/mjt1312/DAFNet}       \\ 
        \hline
        
        MANet &   ICCV-2019  & Multi-Adapter RGBT Tracking &  \url{https://arxiv.org/abs/1907.07485}  & \url{https://github.com/Alexadlu/MANet}       \\ 
        \hline

        \rowcolor[gray]{0.9}  \multicolumn{5}{|c|}{{\textbf{Miscellaneous (RGB+X)  Tracking}}} \\
        \hline 
        
        OneTracker &  CVPR-2024   & OneTracker: Unifying Visual Object Tracking with Foundation Models and Efficient Tuning &  \url{https://arxiv.org/abs/2403.09634}  & \url{}       \\ 
        \hline

        SDSTrack &  CVPR-2024   & SDSTrack: Self-Distillation Symmetric Adapter Learning for Multi-Modal Visual Object Tracking &  \url{https://arxiv.org/abs/2403.16002}  & \url{https://github.com/hoqolo/SDSTrack}       \\ 
        \hline
              
        Un-Track &  CVPR-2024   & Single-Model and Any-Modality for Video Object Tracking &  \url{https://arxiv.org/abs/2311.15851}  & \url{https://github.com/Zongwei97/UnTrack}       \\ 
        \hline
        
        ELTrack &  ArXiv-2024 & ELTrack: Correlating Events and Language for Visual Tracking &  \url{https://papers.ssrn.com/sol3/papers.cfm?abstract_id=4764503}  & \url{https://github.com/HamadYA/ELTrack-Correlating-Events-and-Language-for-Visual-Tracking}       \\ 
        \hline
        
        KSTrack &  TCSVT-2024   & Knowledge Synergy Learning for Multi-Modal Tracking &  \url{https://ieeexplore.ieee.org/document/10388341}  & \url{}       \\ 
        \hline

        SeqTrackv2 &  ArXiv-2024  & Unified Sequence-to-Sequence Learning for Single- and Multi-Modal Visual Object Tracking &  \url{https://arxiv.org/abs/2304.14394}  & \url{https://github.com/chenxin-dlut/SeqTrackv2}       \\ 
        \hline
              
        ViPT &  CVPR-2023   & Visual Prompt Multi-Modal Tracking &  \url{https://arxiv.org/abs/2303.10826}  & \url{https://github.com/jiawen-zhu/ViPT}       \\ 
        \hline
        
        ProTrack &  ACM MM-2022   & Prompting for Multi-Modal Tracking  &  \url{https://arxiv.org/abs/2207.14571}  & \url{}       \\ 
        \hline
        

    \end{longtable}
\end{center}
}

\end{document}